\documentclass[a4paper]{article}

\usepackage{INTERSPEECH2022}

\title{INTERPRETABLE DYSARTHRIC SPEAKER ADAPTATION BASED ON OPTIMAL-TRANSPORT}
\name{Rosanna Turrisi$^{1,2,3}$, Leonardo Badino$^4$}
\address{
  $^1$University of Genoa,
  $^2$Italian Institute of Technology,
  $^3$University of Ferrara,
  $^4$ PerVoice}
\email{rosanna.turrisi@edu.unige.it, leonardo.badino@pervoice.it}

\begin{document}
%\ninept
%
\maketitle
\begin{abstract}
This work addresses the mismatch problem between the distribution of training data (\textit{source}) and testing data (\textit{target}), in the challenging context of dysarthric speech recognition. We focus on Speaker Adaptation (SA) in command speech recognition, where data from multiple sources (i.e., multiple speakers)  are available. Specifically, we propose an unsupervised Multi-Source Domain Adaptation (MSDA) algorithm based on optimal-transport, called MSDA via Weighted Joint Optimal Transport (MSDA-WJDOT). We achieve a Command Error Rate relative reduction of 16$\%$ and 7$\%$ over the speaker-independent model and the best competitor method, respectively. 
The strength of the proposed approach is that, differently from any other existing SA method, it offers an interpretable model that can also be exploited, in this context, to diagnose dysarthria without any specific training. Indeed, it provides a closeness measure between the target and the source speakers, reflecting their similarity in terms of speech characteristics. Based on the similarity between the target speaker and the healthy/dysarthric source speakers, we then define the healthy/dysarthric score of the target speaker that we leverage to perform dysarthria detection. This approach does not require any additional training and achieves a 95$\%$  accuracy in the dysarthria diagnosis. 

\end{abstract}
\section{Introduction}
\label{sec:intro}

%Many machine learning algorithms assume that the test and training datasets are
%sampled from the same distribution. However, in many real-world applications, new
%data can exhibit a distribution change (\textit{domain shift}) that degrades the algorithm
%performance. %In speech recognition, this shift can be observed when training and testing conditions are different (e.g., different recording conditions). 
%This problem can be addressed through Domain Adaptation (DA) \cite{Jiang2007, Wouter2019}, that is a particular case of transfer learning \cite{Pan2009}. DA methods consist in leveraging a labelled dataset, called \textit{source} domain, to build a model that performs well on the dataset of interest, called \textit{target} domain. Source and target domains are supposed to be similar and, typically, the labels for the target domain are not available (unsupervised DA). 

%In speech recognition, a domain shift can be caused by differences between speakers (e.g., different accents, different speaking style). This mismatch is even more evident in speech-impaired people as the speech characteristics also depend on the type and the severity of the disorder.% We refer to this problem as \textit{speaker adaptation}.
%\\

Many Automatic Speech Recognition (ASR) models assume that test and training datasets are sampled from the same distribution. However, new data can exhibit a distribution change and drastically decrease the recognition performance. This is due, for example, to different characteristics between speakers (e.g., different accents, different speaking style). Such a mismatch is even more evident in speech-impaired people as the speech characteristics also depend on the type and the severity of the disorder. 
This problem can be addressed through supervised speaker adaptation (SA), where transcribed utterances of the target speaker are used to adapt a speaker-independent (SI) model to better recognize the speech of the target speaker.     
%which consists in leveraging a labelled dataset containing several speakers data (called \textit{source} speakers), to build a model that performs well on the speaker of interest (called \textit{target} speaker). 

Here we consider the more frequent unsupervised SA scenario, where the transcriptions of the adaptation utterances are not available. Specifically, we focus on unsupervised SA for spoken command recognition of dysarthric speech. Dysarthria is a motor speech disorder consisting in the disruption of the normal control of the vocal tract musculature, common in conditions such as Parkinson Disease (PD), Amyotrophic Lateral Sclerosis (ALS) and post-stroke motor impairments \cite{mcneil2009clinical, darley1969}. Dysarthric patients tend to have 
mumbled speech, acceleration or deceleration in the speaking rate, abnormal pitch and rhythm. 
However, dysarthric speech is characterized by high variability as the impairment varies based on the location and severity of the neurological damage. This, beside the limited amount of available data, makes dysarthric speech recognition very challenging. Indeed, although many studies improved the ASR performance on dysarthric speech, 
they still poorly perform on low intelligible speech \cite{Espana2016, Vachhani2017, Joy2018, xie2021variational}. That points out the necessity of \textit{ad hoc} SA techniques for dysarthric speakers.

In \cite{mustafa2014}, authors suggest that different adaptation techniques should be adopted based on the impairment severity of the speaker. They employ two large datasets to train two different SI models. The first dataset only includes healthy speakers, while the second dataset contains additional recordings from dysarthric speakers. Adaptation is then performed to better recognize the utterances from a dysarthric corpus. Results show that adaptation methods for dysarthric patients with mild or low dysarthria are more effective when the initial model is trained on healthy speech, whereas the adaptation of severe dysarthric patients improves when the SI model is trained on impaired-speech. Following this direction, \cite{takashima2020two} introduces a two-stage adaptation approach in which a SI model is first adapted to the general speaking style of multiple dysarthric speakers, and then the adapted model is further adapted to the speaker of interest. However, the described techniques require a large amount of data for fine-tuning, that is usually not available. To overcome this problem, \cite{shor2019} proposes to only fine-tune a subset of the network layers to better adapt an ASR model to the dysarthric speech. %\cite{morales2009}, rather than adapting the acoustic models, models the phonetic-articulatory errors made by the target speaker and attempts to correct by estimating  the correct word sequence them by two possible strategies that incorporate the language model and find the best estimate of the correct word sequence. 
%Differently,\cite{morales2009} directly models the phonetic-articulatory errors made by the target speakers by incorporating the language model and estimating the correct corresponding word sequence. 
Differently, \cite{morales2009} adopts confusion matrix transducers within the weighted finite state transducer decoding paradigm, to handle phonetic-articulatory errors made by the target speakers. However, all the aforementioned studies are limited to the supervised framework, whereas the unsupervised speaker adaptation is the most common scenario.

We adopt an unsupervised SA approach based on Optimal Transport (OT) \cite{Monge, Kantorovich}, named Multi-Source Domain Adaptation Weighted Joint Distribution Optimal Transport (MSDA-WJDOT), firstly introduced in \cite{turrisi2020multi} for other applications. In this work, we assume to have access to spoken commands of several speakers (called \textit{source}) and aim at finding a classification function for spoken commands of a new speaker (called \textit{target}) through a two-step procedure. The first step learns a global feature extractor from a SI model, e.g., the last hidden layer of a neural network. The second step learns a target-specific classifier, from the feature space to the output space (i.e., the commands to classify). Specifically, we define i) a weighted combination (with weights  $\pmb{\alpha}$) of the joint probability distributions of the source speakers on the feature-output product space; ii) a proxy joint probability distribution of the target speaker on the feature-output product space, in which the output space is represented by the predictions of the target classifier. The algorithm is trained to jointly look for the optimal weights $\pmb{\alpha}$ and the optimal target classifier that minimize the Wasserstein distance between the distributions defined in i) and ii). 
Intuitively, at each iteration, this method simultaneously selects the sources most similar to the target (by attributing them higher weights) and updates the target classifier to increase this similarity. The advantage of MSDA-WJDOT over the other adaptation approaches is that only the relevant source speakers are used, avoiding redundant or misleading information. Mostly important, the proposed approach offers an interpretable model as we found the source weights $\pmb{\alpha}$ reflecting a similarity between the source speakers and the target speaker in terms of speech characteristics.  Contrary to  \cite{mustafa2014} in which the sources are manually chosen based on the impairment severity of the target speaker, MSDA-WJDOT is able to perform this selection automatically considering also other important similarity aspects (e.g., the gender).

%The proposed method requires two steps: i) training a SI model, composed by a feature extractor and a classifier, on the source training speakers; ii) learning a classifier on the feature space for the target speaker, based on the most similar source speakers.\\
%The latter phase looks for a convex combination of the joint source distributions, on the feature-output product space, with minimal Wasserstein distance to the proxy joint target distribution, in which the labels are replaced by the prediction of a classifier. The source weights and the target classifier are simultaneously learned.  The advantage of MSDA-WJDOT over other MSDA approaches is that the source weights provide a similarity measure between each source and the target domain. This can be leveraged to only select the relevant sources and additionally offers an interpretable model. Indeed, these weights turned out to reflect the similarity between speakers.  

\section{MSDA-WJDOT} \label{sec:WJDOT}

In this section, we recall MSDA-WJDOT, firstly proposed in \cite{turrisi2020multi}. %This method approaches the adaptation problem by estimating the similarity between source and target domains and learning a target classifier using only the most similar sources. Specifically, the similarity between domains is computed in terms of the distance between their probability distributions, by using the Wasserstein distance \cite{Monge, genevay2017learning}. 
In the classical machine learning problem, the aim is to learn a feature extractor and a classifier, such that the combination of these two functions describes the input-output relationship. MSDA-WJDOT assumes to have access to the feature extraction function $g$ from the input space to an embedding space $\mathcal{G}$ and looks for a target classifier defined on $\mathcal{G}$. This is the case in which, for example, a pre-trained network is used to extract meaningful features. Once $g$ is known, MSDA-WJDOT performs adaptation on the feature-output product space. 
However, in the considered framework the feature extractor is not available and, hence, the proposed method becomes a two-step procedure, as described in the following.

\subsection{Global feature extractor}
The first step consists in estimating a global feature extractor, e.g., a function that extracts features from the audio of a \textit{general} speaker. To do that, we learn a SI model, composed by a feature extraction function $g$ and a classifier $f_{SI}$. Let us suppose to have $N_{j}$ utterances $(X_{j}, Y_{j})$ of $J$ source speakers, representing the utterance acoustic features $X_{j}$ and the corresponding command label $Y_{j}$. We can train the model on the source speakers $\{(X_{j}, Y_{j})\}_{j=1}^{J}$ by 
\begin{equation}\label{eq:baseline}
 \operatorname*{min}_{g, f_{SI}} \quad\sum _{j=1}^{J} \frac{1}{N_{j}} \sum _{i=1}^{N_{j}} \mathcal{L}(f_{SI} \circ g(x_{j}^{i}), y_{j}^{i}),
\end{equation} 
where $\mathcal{L}$ is the cross-entropy loss function. Note that in the second step, $f_{SI}$ will not be used. 

\subsection{Adaptation}
Once $g$ is given, we can define the joint distributions $p_{S_{j}}$, for each source speaker $j$, with support on the product space $\mathcal{G}\times\mathcal{Y}$, where $\mathcal{Y}$ is the label space. We then define a convex combination of the source distributions
$$    p_{S}^{\alpha} = \sum _{j=1}^{J} \alpha _{j}p_{S_{j}}$$
with $\pmb{\alpha}$ element of the simplex $\Delta^{J}$ (i.e., $0\leq\alpha_{j}\leq1$ for each $j$ and $\sum _{j=1}^{J} \alpha_{j}=1$).
Finally, we define the proxy target joint distribution $p_T^{f}$, with support on
 $\mathcal{G}\times f(\mathcal{G})$, where $f$ is the target classifier to learn. Here, the upper index $f$ indicates that the joint distribution is over predicted labels and not reference labels. MSDA-WJDOT aims at finding the $\pmb{\alpha}$ weights and the classifier $f$ that align the source combination of distributions $p_{S}^{\alpha}$ and the proxy target distribution $p_T^{f}$ . This can be expressed as 
\begin{equation}\label{eq:DA1}
 %\pmb{\alpha}^{*}, f^{*} = 
 \operatorname*{min}_{\pmb{\alpha}, f} \quad
W_D\left( p_T^{f}, p_{S}^{\alpha}\right) = \operatorname*{min}_{\pmb{\alpha}, f} \quad
W_D\left( p_T^{f}, \sum _{j=1}^{J} \alpha _{j} p_{S,j}\right),
\end{equation}
where $W_{D}$ is the Wasserstein distance \cite{Monge, genevay2017learning}. Roughly speaking, the source joint distributions can be seen as proxy joint distributions with support on 
$\mathcal{G}\times f_{j}(\mathcal{G})$, where $f_{j}$ is the true classifier (i.e., $f_{j}(X_{j})\equiv Y_{j}$). This approach dynamically selects the most similar sources and re-adapts $f$ by  ``imitating" the best $f_{j}$.

\subsection{The role of $\pmb{\alpha}$ and the similarity}
We can consider $\pmb{\alpha}$ as a measure of similarity between the sources and the target, in terms of joint probability distribution. Indeed, in order to minimize Eq. \ref{eq:DA1}, the algorithm will attribute higher weights to the source distributions closest to the proxy target distribution, and smaller (or zero) weights to far source distributions. The zero-weighted distributions will not contribute to the adaptation procedure. Consequently, misleading and unrelated source domains are in practice not used and this allows to learn a more accurate classifier. For example, if $\pmb{\alpha}$ were a one-hot vector with $\alpha _{j}=1$ , the method would consist in estimating $f$ such that $p_{T}^{f}\simeq p_{S_{j}}$. Authors in \cite{turrisi2020multi} only offer a mathematical explanation of the weight alpha. In this study, we explore the interpretation of alpha in terms of speaker similarity and speech characteristics.

%\subsection{A practical insight}
%Given the dataset $(x_{j}^{i}, y_{j}^{i})_{i}$ of the source speaker $j$, the empirical joint distribution can be estimated by
%$$p_{S,j} = \frac{1}{N_{j}} \sum _{i=1}^{N_{j}} \delta _{g(x_{j}^{i}, y_{j}^{i})},$$
%where $\delta$ is the Delta Dirac function.  Similarly, for the target speaker, the proxy joint distribution is computed by replacing the label with the classifier prediction  
%$$p_T^f = \frac{1}{N_{T}} \sum _{i=1}^{N_{T}} \delta _{g(x^{i}, f(g(x^{i})))}.$$  

\section{Experimental setup} \label{sec:Exp}

\subsection{Dataset} 
We employ the AllSpeak dataset \cite{dinardi}, that consists of speech recordings from 29 Italian native speakers. Seventeen of these (thirteen males, four females) are affected by Amyotrophic Lateral Sclerosis, while the remaining twelve (six males, six females) are healthy control speakers.  The dataset contains 25 commands in Italian, relative to basic needs such as ``I am thirsty", ``I am hungry". This dataset is very challenging due to the small amount of recordings. Indeed, only 2387 and 1857 examples have been recorded from control and dysarthric speakers, respectively. Specifically, each command has been recorded from 8 to 10 times by healthy speakers, while from 4 to 10 times by patients – depending on the patient medical condition. In the following we will adopt a speaker coding according to which F/M stands for female/male, and the letter C is added to refer to control subjects.

We perform speaker adaptation of each dysarthric speaker (target speaker) by using all the remaining speakers as source speakers. We trained the SI model on the source speakers, by splitting the dataset into training and validation sets. Specifically, the validation set consists of one example of each command of each speaker and it has been used to perform early stopping. The unlabelled target speaker data is split into adaptation set (80$\%$) and testing (20$\%$) set, which contains one example of each command. To train MSDA-WJDOT, we simultaneously employ the source training dataset and the adaptation dataset of the target speaker. 
%Please not that, as the labels are not available for the target speaker, the standard early stopping in MSDA-WJDOT cannot be applied. To overcome this problem, we use the sum of squared errors (SSE). More precisely, we estimate the cluster membership of target point $g(x)$ on the output space (by computing $f(g(x))$) and measure the SSE between $g(x)$ and its cluster centroid. Intuitively, if the SSE decreases it means that $f$ attributes the same label to samples of the target speaker that are close in the feature space $\mathcal{G}$. 

\subsection{Model details}
The feature extraction function $g$ is represented by a BLSTM with five hidden layers, each containing 250 memory blocks. Both the SI classifier $f_{SI}$ and the target classifier $f$ are represented by a softmax layer. All weights are initialized with Xavier initialization. Both SI model training and adaptation are performed with Adam optimizer with 0.9 momentum and $\epsilon = e^{-8}$. Learning rate is fixed to 0.001.

\section{Results} \label{sec:results}

\subsection{Competitor models}
We compare the proposed approach with an unadapted speaker- independent (SI) model, in which the target speaker is directly tested on a model trained on all source speakers. 
Notice that the literature lacks unsupervised MSDA specific for dysarthric speaker adaptation. Hence, we implemented two MSDA-WJDOT approaches - CJDOT and MJDOT (details can be found in \cite{turrisi2020multi}) - which are different extensions of a state-of-the-art unsupervised domain adaptation method - JDOT \cite{Courty2017} - adapted to the multi-speaker adaptation case. In addition, we trained a supervised SA (SSA) model, in which  we added a feed-forward linear layer atop the input to the SI model and trained it on the target speaker \cite{neto1995}. As this method uses the target labels, it must be considered as the lower bound of the unsupervised SA performance error. 
\\

\subsection{Command speech recognition}\label{sec:CSR}
%\begin{table}[th]
%\caption{Command Error Rate (CER) for each dysarthric target speaker provided by the SI, MSDA-WJDOT and the competitor models.}
%\begin{center}
%\resizebox{\linewidth}{!}{% 
%\begin{tabular}{c|cccc} 
%\textbf{Speaker} &SI & CJDOT \cite{Courty2017} & MJDOT \cite{Courty2017} & MSDA-WJDOT\\
%\hline
%\textbf{M01} & 35.79 & 32.77 & \textbf{31.93} & \textbf{31.93} \\
%\textbf{M02}& \textbf{34.26} & 36.75 & 36.75 & 37.71\\ 
%\textbf{F01}& 63.16 & 52.63 & \textbf{49.12} &  \textbf{49.12} \\ 
%\textbf{F02}& 48.50 & \textbf{40.00} & \textbf{40.00} & \textbf{40.00} \\ 
%\textbf{M03} & 64.44 & 68.89 & 71.11 & \textbf{57.78}\\ 
%\textbf{M04} & \textbf{30.00} & 31.20 & 32.00 & 30.40 \\ 
%\textbf{M05} & 18.62 & 17.46 & 15.08 & \textbf{14.29} \\ 
%\textbf{F03} & 68.33 & \textbf{61.74} & 70.43 & 62.61 \\ 
%\textbf{M06} & 48.67 & 34.78 & \textbf{33.91} & 35.65\\ 
%\textbf{M07} & 11.00 & \textbf{7.20} & 8.00 &  8.80 \\ 
%\textbf{M08} & 39.50 & 36.00 & 41.60 & \textbf{33.60} \\ 
%\textbf{M09} & 24.79 & \textbf{16.81} & 18.49 & 19.33 \\ 
%\textbf{F04} & 48.07 & \textbf{38.60} & \textbf{38.60} & \textbf{38.60} \\ 
%\textbf{M10} & 18.00 & 12.80 & \textbf{12.00} & 12.80 \\ 
%\textbf{M11} & 56.50 & 47.20 & 48.80 & \textbf{45.60} \\ 
%\textbf{M12} & 7.50 & 5.60 & 7.20 & \textbf{4.80} \\ 
%\textbf{M13} & 30.91 & 45.45 & 43.64 & \textbf{21.82} \\ 
%\hline
%\textbf{A. CER} & 38.11& 34.46 & 34.37 & \textbf{32.05}\\
%\textbf{A. Rank} & 2.88 & 1.94 & 2.29  & \textbf{1.65} \\
%\end{tabular}}
%\end{center}
%\label{tab:SA}
%\end{table} 

\begin{table}[t]
\caption{Command Error Rate (CER) for each dysarthric target speaker provided by the SI, SSA,  MSDA-WJDOT and the competitor models.}
\begin{center}
\resizebox{\linewidth}{!}{% 
\begin{tabular}{c||c|ccc|c} 
\textbf{Speaker} &SI & CJDOT \cite{Courty2017} & MJDOT \cite{Courty2017} & MSDA-WJDOT & SSA \cite{neto1995}\\
\hline
\textbf{M01} & 35.79 & 32.77 & \textbf{31.93} & \textbf{31.93} & 25.00 \\
\textbf{M02}& 34.26 & \textbf{36.75} & \textbf{36.75} & 37.71 & 54.17\\ 
\textbf{F01}& 63.16 & 52.63 & \textbf{49.12} &  \textbf{49.12} & 60.00 \\ 
\textbf{F02}& 48.50 & \textbf{40.00} & \textbf{40.00} & \textbf{40.00} & 36.00 \\ 
\textbf{M03} & 64.44 & 68.89 & 71.11 & \textbf{57.78} & 55.56 \\ 
\textbf{M04} & 30.00 & 31.20 & 32.00 & \textbf{30.40} & 24.00\\ 
\textbf{M05} & 18.62 & 17.46 & 15.08 & \textbf{14.29} & 17.39 \\ 
\textbf{F03} & 68.33 & \textbf{61.74} & 70.43 & 62.61 & 56.52 \\ 
\textbf{M06} & 48.67 & 34.78 & \textbf{33.91} & 35.65 & 26.09 \\ 
\textbf{M07} & 11.00 & \textbf{7.20} & 8.00 &  8.80  & 20.00 \\ 
\textbf{M08} & 39.50 & 36.00 & 41.60 & \textbf{33.60} & 32.00 \\ 
\textbf{M09} & 24.79 & \textbf{16.81} & 18.49 & 19.33 & 13.04 \\ 
\textbf{F04} & 48.07 & \textbf{38.60} & \textbf{38.60} & \textbf{38.60} & 40.91 \\ 
\textbf{M10} & 18.00 & 12.80 & \textbf{12.00} & 12.80 & 16.00 \\ 
\textbf{M11} & 56.50 & 47.20 & 48.80 & \textbf{45.60} & 40.00\\ 
\textbf{M12} & 7.50 & 5.60 & 7.20 & \textbf{4.80} &4.00 \\ 
\textbf{M13} & 30.91 & 45.45 & 43.64 & \textbf{21.82} & 9.09\\ 
\hline
\textbf{A. CER} & 38.11& 34.46 & 34.37 & \textbf{32.05} & 31.16\\
\end{tabular}}
\end{center}
\label{tab:SA}
\end{table}

Table \ref{tab:SA} reports the results in terms of Command Error Rate (CER). A first remark is that, although the SI model always achieved a CER between $15\%$ and $20\%$ on the source validation set, it often had low accuracy on the target speaker. Once again, this emphasizes the difficulty of an ASR system to generalize to a new dysarthric speaker and the importance of the speaker adaptation in this context.  \\
The unsupervised SA carried out by MSDA- WJDOT outperforms all the methods by providing the best Average CER. Indeed, it reduces the CER of $16\%$ and $7\%$ over the SI system and the best competitor model, respectively. Surprisingly, MSDA-WDJOT achieves an Average CER similar to the SSA approach, in which the labels are used. \\
%Further, we provide the Average Rank that is a performance measure suitable when several targets domain are tested. To every method, it assigns a rank (from 1 to 4, that is the number of considered methods) for each tested target based on the CER (e.g., 1 if the method has the lowest CER, 4 for the highest CER) and then it computes the average of the ranks. This measure is more robust to the variance and confirms that MSDA-WJDOT provides the best performance. \\

\begin{figure}[t]
\centering\includegraphics[width=.9\linewidth]{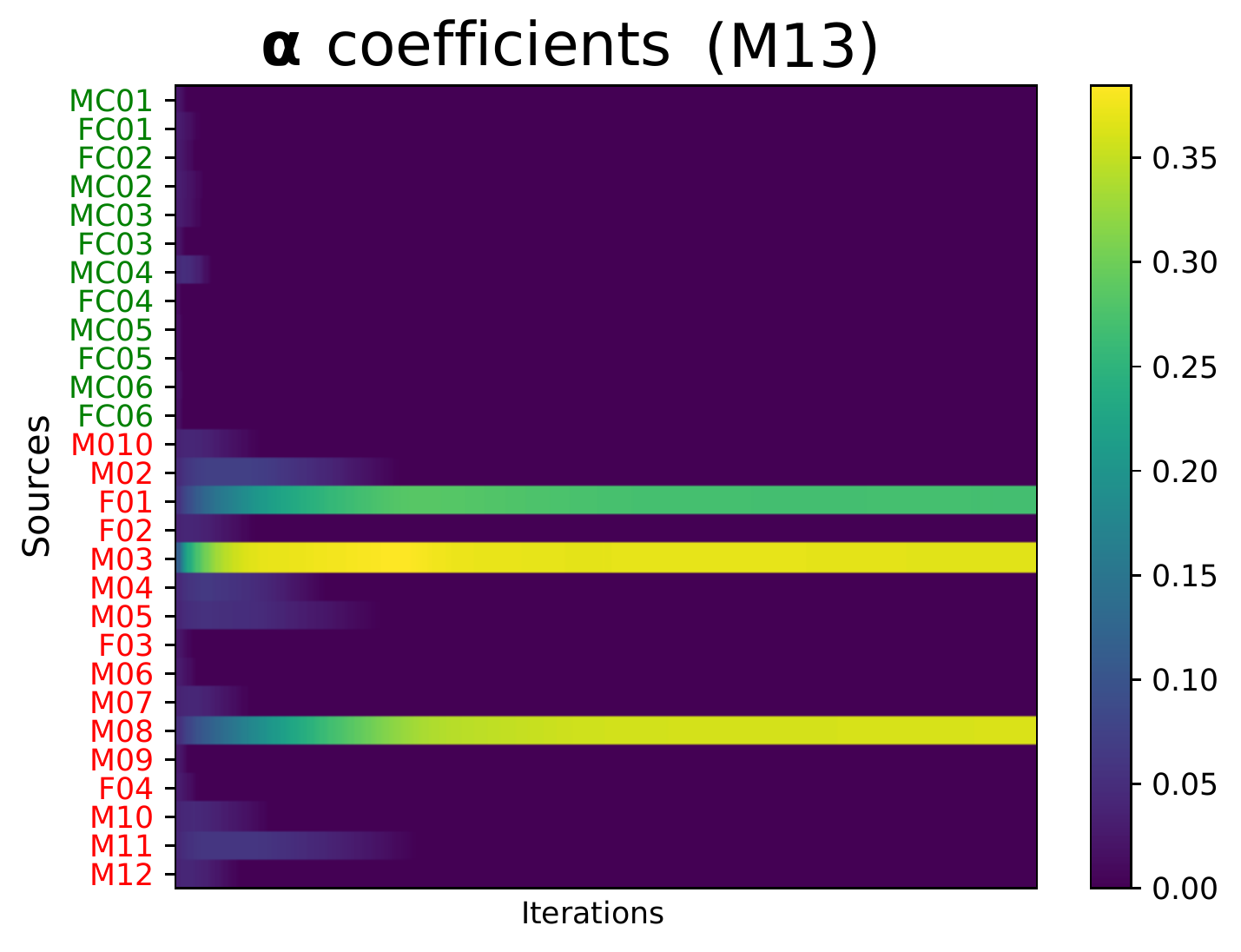}
\caption{$\pmb{\alpha}$ coefficients recovered during MSDA-WJDOT training for dysarthric target speaker M13. The $\alpha _{j}$ coefficients are close to zero for healthy speakers (in green), while the highest weights are attributed to dysarthric speakers (in red).}
\label{fig:alpha_M13}
\end{figure}

It is essential to recall that the proposed approach provides a measure of similarity between the source speakers and the target speaker, in terms of probability distribution. We found that the recovered $\pmb{\alpha}$ always attributes highest similarity scores to dysarthric speakers rather than healthy ones, when the target speaker is dysarthric, and vice versa for healthy target speakers. This means that $\pmb{\alpha}$ weights can catch the similarity in terms of speech characteristics and that this approach can realistically estimate speakers similarity.
Fig. \ref{fig:alpha_M13} shows the recovered weights for a dysarthric target speaker (M13) during the training. Specifically, we report on the $x$-axis the epochs of training, while the $y$-axis lists the healthy (in green) and dysarthric (in red) source speakers. As we can see, after few epochs $\alpha_{j}$ becomes sparse, using in practice only the data of the most similar speakers. Interestingly, all the healthy speakers have zero weights whereas the $\pmb{\alpha}$ mass is spread along dysarthric speakers. Also, the highest weights are attributed to M03 and M08, which are both males, while a lower importance is attributed to F01. This suggests that MSDA-WJDOT also incorporates similarity information about the gender.

\begin{figure}[t]
\centering\includegraphics[width=.9\linewidth]{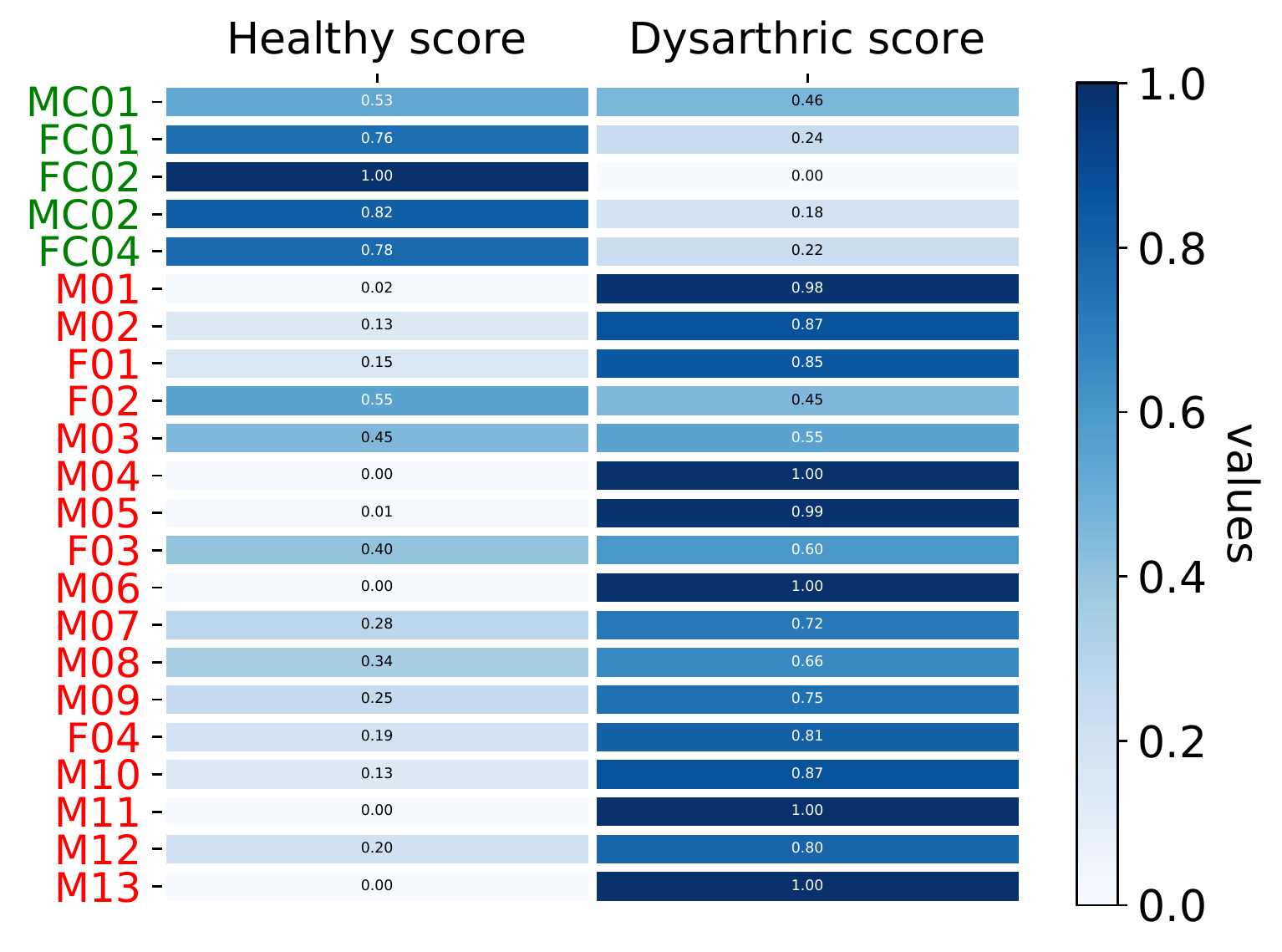}
\caption{HS and DS computed for healthy (in green) and dysarthric (in red) speakers.}
\label{fig:alpha_all_scores}
\end{figure}

\subsection{The $\pmb{\alpha}$ weight and the dysarthria detection}
We here propose a completely novel approach to perform dysarthia detection based on the $\pmb{\alpha}$ weights. As shown in Fig. \ref{fig:alpha_M13}, MSDA-WJDOT associates speakers with similar voice characteristics. We investigated the possibility of leveraging the $\pmb{\alpha}$ weights of the command classifier to infer if the target speaker is affected by dysarthria. Specifically, we propose to classify a speaker as healthy or dysarthric based on his/her similarity with the other subjects.

Let define $I_{c}$ as the set indexing the control speakers and $I_{d}$ as the set of indices related to dysarthric speakers. We then define the Healthy Score (HS) and the Dysarthric Score (DS) as follow 
\begin{equation}
    HS = \sum _{j\in I_{c}} \alpha _{j}, \quad DS = \sum _{j\in I_{d}} \alpha _{j}.
\end{equation}
As $\pmb{\alpha}\in\Delta^{J}$, we have that $HS, DS\in[0, 1]$ such that $HS + DS = 1$. Fig.  \ref{fig:alpha_all_scores} reports the computed scores for all dysarthric speakers and for 5 control speakers. The control subjects always show a higher $HS$, while for the patients (except for F02), we have a higher $DS$. Moving a step forward, we can use these scores to perform dysarthria detection by stating that
\begin{center}
\textit{A speaker is affected by dysarthria if} $DS > HS$.
\end{center}
%$$A\; speaker\; is\; affected\; by\; dysarthria\; if\; DS > HS.$$
This results in a final accuracy of 95$\%$. It is crucial to emphasize that, differently from standard dysarthria detection approaches \cite{williamson2015, tu2017, hernandez2020}, any specific training has been performed: we directly infer the speaker health status from the similarity coefficients $\pmb{\alpha}$ that provide us an interpretable model. Note that none of existing SA methods are able to also perform dysarthria detection. Hence, we cannot fairly compare our approach to any competitor model.

% \begin{figure*}
%   \begin{minipage}{.45\textwidth}
% \centering\includegraphics[width=.9\linewidth]{alpha_M13.pdf}
% \caption{$\pmb{\alpha}$ coefficients recovered during MSDA-WJDOT training for dysarthric target speaker M13. The $\alpha _{j}$ coefficients are close to zero for healthy speakers (in green), while the highest weights are attributed to dysarthric speakers (in red).}
% \label{fig:alpha_M13}
%   \end{minipage} \quad
%   \begin{minipage}{.45\textwidth}
% \centering\includegraphics[width=.9\linewidth]{alpha_all_scores.pdf}
% \caption{HS and DS computed for healthy (in green) and dysarthric (in red) speakers.}
% \label{fig:alpha_all_scores}
%   \end{minipage}
% \end{figure*}

\section{Conclusions}\label{sec:discussion}
In this work, we addressed usupervised speaker adaptation in the challenging context of pathological speech by employing MSDA-WJDOT algorithm. Our approach significantly improves the speaker-independent system, reducing the CER of $16\%$, and it also outperforms all the competitor models. The advantage of the proposed method is that it does not require the labels for the target speaker and, interestingly, it  also provides source-target probability distribution similarity coefficients $\pmb{\alpha}$, reflecting the source-target speakers characteristics similarity. From these coefficients, we derived the Healthy (HI) and Dysarthric (DI) Index of a target speaker that allowed us to detect dysarthria with an accuracy of 95$\%$.

The novelty of our work relies on several aspects.
Firstly, this work brings a contribution in the field of dysarthric speaker adaptation, in which unsupervised methods dramatically lack while they are essential for a good recognition accuracy.  
The second type of novelty is the interpretation of the method. To the best of our knowledge, this is the first SA study in which the model provides a measure of similarity between speakers. Note that this approach may be transferred to other SA contexts in which, for instance, the algorithm may exploit speakers accents and provide a similarity in terms of region of origin of the speakers. Last but not least, we moved a step forward and we introduced new ranking indices that we used to perform dysarthria detection. This approach is completely new and can achieve high diagnosis performance, without requiring a specific model training. 

Future directions could delve into the dysarthria assessment by individuating intervals of values in which the DI corresponds to dysarthria severity levels (e.g., mild, moderate, severe).
This may bring to very efficient ASR systems that simultaneously improve their performance via SA, and compute the DI warning the subject when the index is close to the right endpoint of its interval. Such a device could  perform an early detection of the disorder or predict the disease degeneration and allow the patient to act in time in order to prevent it. Furthermore, the dysarthric index may be used to track the speaker changes during a treatment and monitoring the benefits of the rehabilitation.

Finally, our MSDA-WJDOT approach was applied on a limited resource scenario where the initial unadapted speaker independent model was trained on the AllSpeak training set. In future work we will experiment with speaker independent models whose hidden layers are pre-trained on much larger datasets (e.g., datasets used for ASR).


\bibliographystyle{IEEEtran}

\bibliography{mybib}

% \begin{thebibliography}{9}
% \bibitem[1]{Davis80-COP}
%   S.\ B.\ Davis and P.\ Mermelstein,
%   ``Comparison of parametric representation for monosyllabic word recognition in continuously spoken sentences,''
%   \textit{IEEE Transactions on Acoustics, Speech and Signal Processing}, vol.~28, no.~4, pp.~357--366, 1980.
% \bibitem[2]{Rabiner89-ATO}
%   L.\ R.\ Rabiner,
%   ``A tutorial on hidden Markov models and selected applications in speech recognition,''
%   \textit{Proceedings of the IEEE}, vol.~77, no.~2, pp.~257-286, 1989.
% \bibitem[3]{Hastie09-TEO}
%   T.\ Hastie, R.\ Tibshirani, and J.\ Friedman,
%   \textit{The Elements of Statistical Learning -- Data Mining, Inference, and Prediction}.
%   New York: Springer, 2009.
% \bibitem[4]{YourName17-XXX}
%   F.\ Lastname1, F.\ Lastname2, and F.\ Lastname3,
%   ``Title of your INTERSPEECH 2022 publication,''
%   in \textit{Interspeech 2022 -- 23\textsuperscript{rd} Annual Conference of the International Speech Communication Association, September 18-22, Incheon, Korea, Proceedings, Proceedings}, 2022, pp.~100--104.
% \end{thebibliography}

\end{document}